\documentclass[10pt,twocolumn,letterpaper]{article}

\usepackage{iccv}
\usepackage{times}
\usepackage{epsfig}
\usepackage{graphicx}
\usepackage{amsmath}
\usepackage{amssymb}

\usepackage{booktabs}
\usepackage{cite}
\usepackage{caption}
\usepackage{multirow}
\usepackage{multicol}
\usepackage{color}
\usepackage{xcolor}
\definecolor{citecolor}{HTML}{0071bc}

\usepackage[pagebackref=true,breaklinks=true,letterpaper=true,colorlinks,citecolor=citecolor,bookmarks=false,urlcolor=citecolor]{hyperref}

\DeclareMathOperator*{\argmin}{arg\,min}

\newcommand{\tablestyle}[2]{\setlength{\tabcolsep}{#1}\renewcommand{\arraystretch}{#2}\centering\footnotesize}

\newlength\savewidth\newcommand\shline{\noalign{\global\savewidth\arrayrulewidth
  \global\arrayrulewidth 1pt}\hline\noalign{\global\arrayrulewidth\savewidth}}

\iccvfinalcopy %

\ificcvfinal\fi

\begin{document}

\title{FeatureNeRF: Learning Generalizable NeRFs by Distilling Foundation Models}

\author{Jianglong Ye$^{1}$\qquad  Naiyan Wang$^{2}$\qquad Xiaolong Wang$^{1}$ \\
  $^{1}$UC San Diego\qquad $^{2}$TuSimple }

\twocolumn[{%
      \vspace{-1em}
      \maketitle
      \vspace{-1em}
      \begin{center}
        \centering
        \vspace{-0.2in}
        \includegraphics[width=\linewidth]{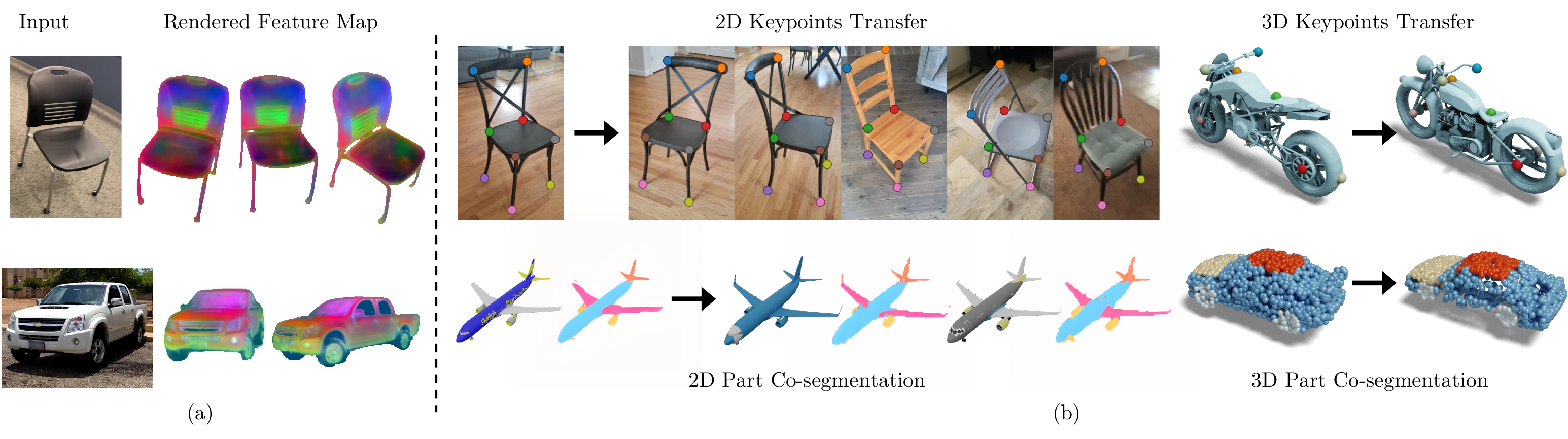}
        \vspace{-0.3in}
        \captionof{figure}{
          While most generalizable NeRFs focus on novel-view synthesis, we propose a framework named FeatureNeRF to learn 3D semantic representations by distilling vision foundation models. After distillation, FeatureNeRF allows to render novel-view feature maps given a single  input image (a), which can be leveraged to various downstream tasks. Here, we show how we propagate part segmentation labels and keypoints to different views and instances in both 2D and 3D domains (b).
        }
        \label{fig:teaser}
      \end{center}
    }]

\maketitle
\ificcvfinal\thispagestyle{empty}\fi

\begin{abstract}
   Recent works on generalizable NeRFs have shown promising results on novel view synthesis from single or few images. However, such models have rarely been applied on other downstream tasks beyond synthesis such as semantic understanding and parsing. In this paper, we propose a novel framework named FeatureNeRF to learn generalizable NeRFs by distilling pre-trained vision foundation models (e.g., DINO, Latent Diffusion). FeatureNeRF leverages 2D pre-trained foundation models to 3D space via neural rendering, and then extract deep features for 3D query points from NeRF MLPs. Consequently, it allows to map 2D images to continuous 3D semantic feature volumes, which can be used for various downstream tasks. We evaluate FeatureNeRF on tasks of 2D/3D semantic keypoint transfer and 2D/3D object part segmentation. Our extensive experiments demonstrate the effectiveness of FeatureNeRF as a generalizable 3D semantic feature extractor. Our project page is available at \url{https://jianglongye.com/featurenerf/}.
\end{abstract}

\section{Introduction}
\label{sec:intro}

Neural fields have emerged as a compelling paradigm for representing a variety of visual signals~\cite{mescheder2019occupancy,park2019deepsdf,chen2019learning,mildenhall2021nerf,chen2021learning}. In particular, the Neural Radiance Fields (NeRF~\cite{mildenhall2021nerf}), which implicitly encodes density and color via Multi-Layer Perceptrons (MLPs), has shown high quality novel view synthesis results from dense input images. A body of follow-up works~\cite{yu2021pixelnerf,chen2021mvsnerf,reizenstein2021common,trevithick2020grf,lin2023visionnerf} further reduce the dependency on dense inputs and generalizes NeRF to unseen objects by learning priors from large-scale multi-view image datasets. With the remarkable abilities on reconstruction and view synthesis of generalizable NeRFs, we ask the question: Can we adapt such models to learn 3D representations as foundations for general 3D applications (e.g., recognition, matching) beyond view synthesis?

Recent years have witnessed the rise of vision foundation models~\cite{yuan2021florence,caron2021emerging,ramesh2021zero,radford2021learning} that are pre-trained on web-scale image datasets and demonstrate generalization capabilities across massive vision tasks (e.g., CLIP~\cite{radford2021learning}, DINO~\cite{caron2021emerging}, Latent Diffusion~\cite{rombach2022high}). The feature space constructed by foundation models captures rich semantic and structural information of 2D visual data and make it possible to identify object categories, parts and correspondences even without extra supervisions~\cite{amir2021deep,caron2021emerging, melas2022deep}. Motivated by these works, our goal is to leverage the powerful 2D foundations models to obtain generalizable 3D features.

In this paper, we present FeatureNeRF, a unified framework for learning generalizable NeRFs from distilling pre-trained 2D vision foundation models. Unlike previous generalizable NeRFs~\cite{yu2021pixelnerf,reizenstein2021common}, which utilize 2D encoder solely for novel view synthesis, FeatureNeRF explores the use of deep features extracted from NeRFs as generalizable 3D visual descriptors. We show that distilling 2D foundation models into 3D space via neural rendering equips the NeRF features with rich semantic information. As a result, FeatureNeRF allows to predict a continuous 3D semantic feature volume from a single or a few images, which can be applied to various downstream tasks such as semantic keypoint transfer and object part co-segmentation. Examples of these applications are shown in Fig.~\ref{fig:teaser}.

Specifically, we adopt an encoder to map 2D images to corresponding 3D NeRF volume similar to previous generalizable NeRFs. Apart from density and color, we propose to extract deep features of the query 3D points from the intermediate layers of NeRF MLP. To enrich semantic information of the NeRF features, we further transfer knowledge from the foundation models to the encoder via neural rendering during training: The rendered feature outputs should be consistent with the feature extracted from the foundation models, which is enforced by a distillation loss.

To evaluate FeatureNeRF, we tackle the tasks of 2D/3D semantic keypoint transfer and object part segmentation. To the best of our knowledge, our work is the first to resolve these 3D semantic understanding tasks without 3D supervision. We validate our framework with two foundation models: (i) DINO~\cite{caron2021emerging}, a self-supervised vision transformer aware of object correspondences, and (ii) Latent Diffusion~\cite{rombach2022high}, a diffusion-based model that achieves state-of-the-art text-to-image generation performance. Our extensive experiments demonstrate the effectiveness of FeatureNeRF as a generalizable 3D semantic feature extractor.

\section{Related Work}
\label{sec:related}

\textbf{Generalizable NeRFs.}
In the past few years, neural fields have gained significant attention and led to rapid progress in representing various visual signals~\cite{mescheder2019occupancy,park2019deepsdf,chen2019learning,mildenhall2021nerf,chen2021learning,karras2021alias,sitzmann2019scene,niemeyer2019occupancy,tewari2022advances,xie2022neural}. In particular, NeRF~\cite{mildenhall2021nerf} achieves photo-realistic results on novel view synthesis by mapping 3D coordinates and 2D viewing directions to density and color via MLPs. However, the original NeRF requires enormous posed images and time-consuming optimization for each single scene. To address these issues, a large number of follow-up methods~\cite{yu2021pixelnerf,jang2021codenerf,wang2021ibrnet,rematas2021sharf,reizenstein2021common,trevithick2020grf,lin2023visionnerf,chen2022transformers} propose to learn generalizable NeRFs from large-scale multi-view image datasets. For example, PixelNeRF~\cite{yu2021pixelnerf} employs an image encoder to condition NeRF on image features, which enables novel views synthesis from a single image and generalizes NeRF to unseen objects. CodeNeRF~\cite{jang2021codenerf} learns to disentangle shape and texture by learning separate embeddings, allowing shape and texture editing by varying the latent codes. Recent work TransINR~\cite{chen2022transformers} is proposed to infer NeRF parameters directly with a vision transformer to overcome the information bottleneck of encoder-decoder architecture. However, most of these works focus only on view synthesis. Our work differs from them by learning general-purpose 3D representations for multiple downstream tasks.

\textbf{Vision Foundation Models.} The term foundation model is introduced in \cite{bommasani2021opportunities} to refer to the model pre-trained from data at scale and capable of generalizing to a wide range of downstream tasks. After demonstrating huge impacts in NLP~\cite{devlin2018bert,brown2020language,raffel2020exploring}, a large family of vision foundation models~\cite{radford2021learning,yuan2021florence,xie2021self,caron2021emerging,ramesh2021zero,rombach2022high,Wei2022MaskedFP,ramesh2022hierarchical} have been proposed and effectively transferred to various vision tasks. For example, the CLIP model  ~\cite{radford2021learning} is trained from large-scale image-text data using contrastive learning, and it is shown to be transferable to multiple tasks in a zero-shot manner. The DINO model~\cite{caron2021emerging} has shown object segment can emerge automatically with only self-supervision, and the learned feature can be applied in a wide range of visual correspondence and recognition tasks~\cite{melas2022deep, Wang2022SelfSupervisedTF,Choudhury2021UnsupervisedPD,amir2021deep}. Besides contrastive learning, recent text-conditioned generative model such as diffusion models~\cite{Ho2020DenoisingDP,Nichol2021ImprovedDD,Dhariwal2021DiffusionMB,rombach2022high} have been introduced and shown astonishing performance on image generation. Subsequently, feature spaces learned by these generative models have also been used for recognition tasks such as semantic segmentation~\cite{Baranchuk2022LabelEfficientSS,Wolleb2021DiffusionMF}. In contrast to the success of 2D foundation models, the 3D counterparts are still suffering from the lack of large-scale annotated datasets and effective architectures~\cite{Wu2021TowersOB,Ha2022SemanticAO,Rozenberszki2022LanguageGroundedI3}. In this paper, we propose to distill the features from 2D foundation models to 3D space via the generalizable NeRFs.

\textbf{Feature Distillation.} For the purpose of model compression and knowledge transfer, distillation has been widely studied by the community. After the pioneering work by Hinton et al.~\cite{Hinton2015DistillingTK}, which matches the softmax output distribution of the teacher model to that of the student, numerous methods have been proposed to tackle various tasks~\cite{Park2019RelationalKD,Heo2019ACO,Passalis2018LearningDR,Tian2020ContrastiveRD,Huang2017LikeWY,Yim2017AGF}. Recently, researchers also propose to distill features from 2D models to 3D space by optimizing neural feature fields~\cite{tschernezki2022neural,kobayashi2022decomposing}. Multiple editing tasks are shown as applications. However, these methods not only require test-time optimization for each single scene, but the learned features are also not generalizable to unseen objects, which makes them unsuitable for general semantic understanding tasks and differ from our work fundamentally.

\begin{figure*}[t]
  \centering
  \vspace{-0.1in}
  \includegraphics[width=1.0\textwidth]{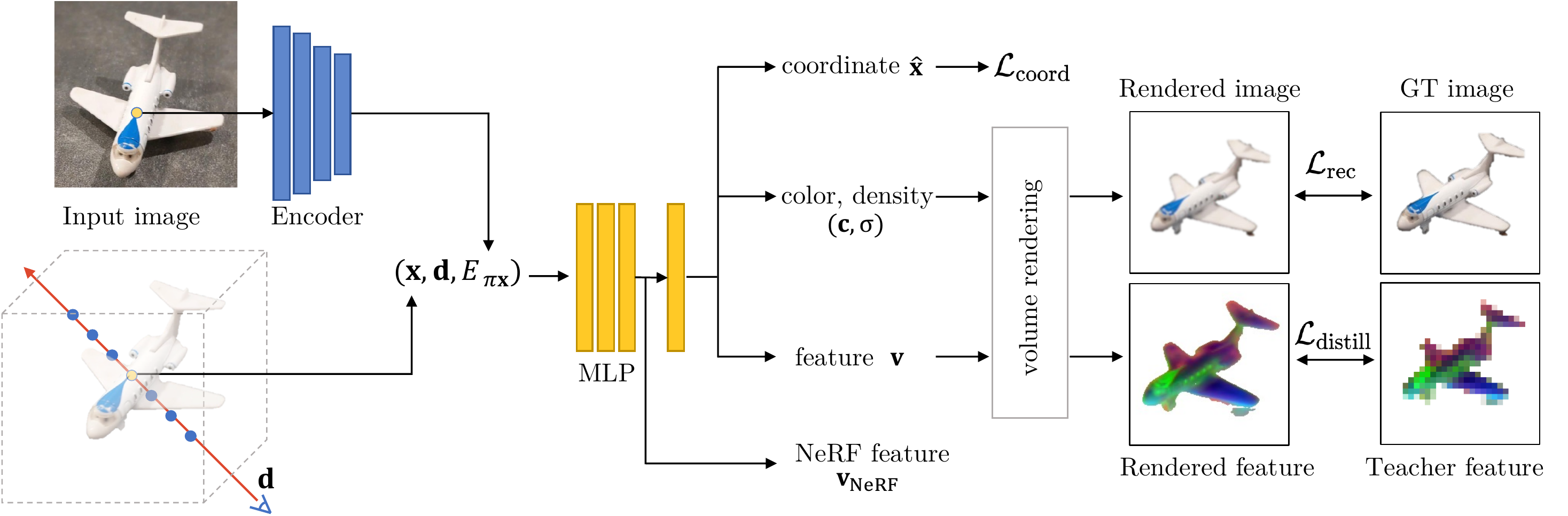}
  \vspace{-0.25in}
  \caption{\textbf{Pipeline of FeatureNeRF}. Given a single image $I$ as input, FeatureNeRF adopts an encoder to extract the image feature $E_{\pi(\mathbf{x})}$, and then concatenate it with the query point $\mathbf{x}$ as well as the view direction $\mathbf{d}$ as the inputs for NeRF MLPs. Apart from density $\sigma$ and color $\mathbf{c}$, we add two MLP branches to predict the feature vector $\mathbf{v}$ and coordinate $\hat{\mathbf{x}}$, which are supervised by two novel loss terms $\mathcal{L}_{\mathrm{distill}}$ and $\mathcal{L}_{\mathrm{coord}}$ respectively. Consequently, we distill knowledge from 2D vision foundation models to FeatureNeRF. Besides, we propose to extract internal NeRF feature $\mathbf{v}_{\mathrm{NeRF}}$ as 3D-consistent feature representation.}
  \vspace{-0.2in}
  \label{fig:method}
\end{figure*}

\textbf{Semantic Correspondences.} Given a pair of visual observation, semantic correspondences learning aims to find corresponding points between them. Several supervised~\cite{ Sun2021LoFTRDL,Jiang2021COTRCT,Choy2019FullyCG} and self-supervised~\cite{amir2021deep,Liu2020LearningIF,cheng2021learning} methods have been proposed to resolve this task in 2D and 3D domain respectively. In particular, Amir et al.~\cite{amir2021deep} show that utilizing pre-trained 2D foundation models in a zero-shot manner can achieve competitive results with supervised methods on semantic correspondences. Cheng et al.~\cite{cheng2021learning} propose to learn point cloud correspondences via self-reconstruction and cross-reconstruction of 3D shapes. To the best of our knowledge, our work is the first to address semantic correspondences in both 2D and 3D space with only 2D observations.

\section{Method}
\label{sec:method}

We present FeatureNeRF, a unified framework for learning generalizable NeRF from vision foundation models. We give an overview of generalizable NeRFs in Sec.~\ref{subsec:preliminary} and elaborate our feature distillation process in Sec.~\ref{subsec:distillation}. Then we introduce how to learn internal NeRF features for 3D semantic understanding in Sec.~\ref{subsec:descriptor} and downstream applications in Sec.~\ref{subsec:applications}. The overall pipeline of FeatureNeRF is illustrated in Fig.~\ref{fig:method}.

\subsection{Preliminary: Generalizable NeRF}
\label{subsec:preliminary}

Neural Radiance Fields (NeRF~\cite{mildenhall2021nerf}) consists of two functions: $\sigma(\mathbf{x}): \mathbb{R}^3 \mapsto \mathbb{R}_{+}$ that maps a 3D point $\mathbf{x}$ to the density $\sigma$ and $\mathbf{c}(\mathbf{x}, \mathbf{d}): \mathbb{R}^{3 \times 3} \mapsto \mathbb{R}^3$ that maps a 3D point as well as a unit viewing direction $\mathbf{d}$ color. The radiance field can be rendered and optimized via differentiable volume rendering~\cite{max1995optical}. Given a pixel's camera ray $\mathbf{r}(t) = \mathbf{o} + t \mathbf{d}$, which is defined by the camera origin $\mathbf{o} \in \mathbb{R}^3$, view direction $\mathbf{d}$ and depth $t$ with bounds $\left[t_{n}, t_{f}\right]$, the estimated color of the ray can be calculated by:
\begin{equation}
  \label{eq:volume}
  \hat{\mathbf{C}}(\mathbf{r})=\int_{t_{n}}^{t_{f}} T(t) \sigma\left(\mathbf{r}(t)\right) \mathbf{c}\left(\mathbf{r}(t), \mathbf{d}\right) \mathrm{d} t,
\end{equation}
where $T(t)=\exp \left(-\int_{t_n}^t \sigma(s) d s\right)$. In practice, the integral is approximated with numerical quadrature by sampling points along the ray. NeRF is optimized to a single scene with multi-view posed images by minimizing the following reconstruction loss:
\begin{equation}
  \label{eq:rec}
  \mathcal{L}_{\mathrm{rec}} =\sum_{\mathbf{r} \in \mathcal{R}}\left\|\mathbf{C}(\mathbf{r})-\hat{\mathbf{C}}(\mathbf{r})\right\|_2^2,
\end{equation}
where $\mathbf{C}(\mathbf{r})$ is the ground truth color of the ray and $\mathcal{R}$ is the set of rays generated from camera poses.

In order to generalize to novel scenes, the NeRF model can be conditioned on the input image $I \in \mathbb{R}^{H \times W \times 3}$:
\begin{equation}
  \begin{aligned}
     & \sigma(\mathbf{x}, I) = g_{\sigma}\left(\mathbf{x}, f(I)_{\pi(\mathbf{x})} \right)                                  \\
     & \mathbf{c}(\mathbf{x}, \mathbf{d}, I) = g_{\mathbf{c}}\left(\mathbf{x}, \mathbf{d}, f(I)_{\pi(\mathbf{x})} \right),
  \end{aligned}
\end{equation}
where $g_{\sigma}$ and $g_{\mathbf{c}}$ are MLPs that predict density and color respectively, $f$ is an image encoder and $\pi$ is the projection function.

As shown in the left part of Fig.~\ref{fig:method}, the image $I$ is firstly passed to an encoder  $f_{\mathrm{enc}}$ (blue blocks in the figure) to obtain a feature map $E = f(I)$. The query point $\mathbf{x}$ is projected onto the image plane using known pose and intrinsics to extract the corresponding feature vector $E_{\pi(\mathbf{x})}$. Then the feature vectors are concatenated with the positional-encoded point $\mathbf{x}$ and direction $\mathbf{d}$, and passed to subsequent MLPs $g_{\sigma}$ and $g_{\mathbf{c}}$ (yellow blocks in the figure) to predict appearance and geometry. When multi-view images are available, feature vectors from different views are aggregated with the average pooling before passing to MLPs.

\subsection{Feature Distillation from Foundation Models}
\label{subsec:distillation}

While most generalizable NeRFs only predict density $\sigma$ and color $\mathbf{c}$, it's possible to extend NeRF to predict other quantities of interests. For example, SemanticNeRF~\cite{zhi2021place} and PanopticNeRF~\cite{fu2022panoptic} propose to add a branch to predict segmentation labels to achieve a 3D-consistent semantic segmentation of a scene. However, these methods require expensive semantic labels during optimization, which is impractical for general cases. In this paper, we aim to transfer knowledge from a pre-trained foundation model $f_{\mathrm{teacher}}$ to our generalizable NeRFs to perform 3D semantic understanding. To this end, we add a branch to output a  high-dimensional feature vector $\mathbf{v} \in \mathbb{R}^D$ for the query point $\mathbf{x}$, where $D$ is the feature channels. Similar to the color rendering (Eq.~\ref{eq:volume}), we can aggregate the feature vectors along a ray as follows:
\begin{equation}
  \begin{aligned}
    \label{eq:feature_rendering}
     & \hat{\mathbf{V}}(\mathbf{r}, I) =\int_{t_{n}}^{t_{f}} T(t) \sigma\left(\mathbf{r}(t), I\right) \mathbf{v}\left(\mathbf{r}(t), \mathbf{d}, I\right) \mathrm{d} t \\
     & \mathbf{v}\left(\mathbf{x}, \mathbf{d}, I\right) = g_{\mathbf{v}}\left(\mathbf{x}, \mathbf{d}, f(I)_{\pi(\mathbf{x})} \right),
  \end{aligned}
\end{equation}
where $g_{\mathbf{v}}$ is the MLP that predicts feture vectors. Note that our model is still conditioned on the image $I$ in the above equation.

We then minimize the difference between rendered pixel feature vector $\hat{\mathbf{V}}$ and the teacher's feature $\mathbf{V} = f_{\mathrm{teacher}}(I)_{\pi(\mathbf{x})}$. In this way, we distill the teacher network into our generalizable NeRFs via neural rendering. We add a distillation loss to penalize the difference:
\begin{equation}
  \label{eq:distill}
  \mathcal{L}_{\mathrm{distill}} =\sum_{\mathbf{r} \in \mathcal{R}}\left\|\mathbf{V}(\mathbf{r})-\hat{\mathbf{V}}(\mathbf{r})\right\|_2^2.
\end{equation}

FeatureNeRF can be trained jointly for image reconstruction and feature distillation by combing two losses (Eq.~\ref{eq:rec} and Eq.~\ref{eq:distill}). We show both color rendering and feature rendering processes in the right part of Fig.~\ref{fig:method}.  We emphasize that, after distillation, FeatureNeRF can obtain a 3D semantic feature function $\mathbf{v}$ in a single forward pass, which can be used for downstream applications. Our experiments show that the feature function $\mathbf{v}$, learned with only 2D observation, contains accurate 3D semantic information.

Our framework can be built on top of any foundation model, and in this work we employ DINO~\cite{caron2021emerging} and Latent Diffusion~\cite{rombach2022high} as teacher networks. DINO is a vision transformer trained with self-distillation, and we simply extract features from the deepest layer as teacher features. Latent Diffusion firstly transforms input image $I$ to the latent space, and utilizes a U-Net~\cite{ronneberger2015u} architecture to estimate the noise for the backward diffusion process. Besides, the denoising module can be conditioned on inputs like text and segmentation maps. We add noise to the original image, condition the pre-trained model with fixed language prompts (e.g. ``Car", ``Chair") and extract features from the intermediate layers of U-Net. Distilling these foundation models that pre-trained on large-scale image datasets brings open-world knowledge to our generalizable 3D representation.

\subsection{Learning Internal NeRF Features for 3D Semantic Understanding}
\label{subsec:descriptor}

Although we have distilled features from foundation models, it is still questionable whether the final feature output is best suitable for 3D semantic understanding. Here we explore using internal NeRF features as view-independent representations for 3D semantic understanding and introduce a new coordinate loss for learning spatial-aware NeRF features.

 While previous works only utilize final outputs of MLPs, we explore whether we can use features from intermediate layers as continuous 3D visual descriptors. Given a input image $I$, we firstly learn a function $\mathbf{v}_{\mathrm{NeRF}}(\mathbf{x}, I)$ to predict NeRF features and utilize several shallow MLPs to predict other quantities:
\begin{equation}
  \begin{aligned}
     & \mathbf{v}_{\mathrm{NeRF}}(\mathbf{x}, I) = g_{\mathrm{NeRF}}\left(\mathbf{x}, f(I)_{\pi(\mathbf{x})} \right)             \\
     & \sigma(\mathbf{x}, I) = g_{\sigma}\left(\mathbf{v}_{\mathrm{NeRF}}(\mathbf{x}, I)\right)                                  \\
     & \mathbf{c}(\mathbf{x}, \mathbf{d}, I) = g_{\mathbf{c}}\left(\mathbf{d}, \mathbf{v}_{\mathrm{NeRF}}(\mathbf{x}, I)\right)  \\
     & \mathbf{v}(\mathbf{x}, \mathbf{d}, I) = g_{\mathbf{v}}\left(\mathbf{d}, \mathbf{v}_{\mathrm{NeRF}}(\mathbf{x}, I)\right),
  \end{aligned}
\end{equation}
where $g_{\mathrm{NeRF}}$ is MLP that predicts NeRF features. Note that the proposed function $\mathbf{v}_{\mathrm{NeRF}}$ can be learned without the feature distillation introduced in Sec.~\ref{subsec:distillation}, therefore it can be applied to all generalizable NeRFs. The feature extraction process is demonstrated in the bottom of Fig.~\ref{fig:method}. We compare the performance of NeRF features with and without feature distillation in our experiments.

Even for the feature distillation version, since teacher's features $\mathbf{V}$ are always not 3D-consistent, using a view-independent representation $\mathbf{v}_{\mathrm{NeRF}}$ and modeling view-dependent effect using another MLP $g_{\mathbf{v}}$ that conditioned on view direction $\mathbf{d}$ further boosts the performance (See Sec.~\ref{subsec:ablation} for the ablation study).

The supervision of most generalizable NeRFs are RGB values, which do not contain spatial information. To enhance the spatial perception of the NeRF feature, we propose to utilize another MLP branch $g_{\mathbf{x}}$ (shown at the top of Fig.~\ref{fig:method}) to regress the input coordinates $\mathbf{x}$ given the NeRF feature $\mathbf{v}_{\mathrm{NeRF}}$:
\begin{equation}
  \hat{\mathbf{x}} = g_{\mathbf{x}}\left(\mathbf{v}_{\mathrm{NeRF}}(\mathbf{x}, I) \right).
\end{equation}
We add a cycle-consistent loss to penalize the difference:
\begin{equation}
  \mathcal{L}_{\mathrm{coord}} =\sum_{\mathbf{r} \in \mathcal{R}} \sum_{t = t_{n}}^{t_f} \left\|\mathbf{r}(t)- g_{\mathbf{x}}\left(\mathbf{v}_{\mathrm{NeRF}}\left(\mathbf{r}(t), I\right) \right)\right\|_2^2.
\end{equation}
The final loss function is the weighted sum of all three losses: $\mathcal{L} = \mathcal{L}_{\mathrm{rec}} + \lambda_{\mathrm{distill}} \mathcal{L}_{\mathrm{distill}} + \lambda_{\mathrm{coord}} \mathcal{L}_{\mathrm{coord}}$, where $\lambda_{\mathrm{distill}}$ and $\lambda_{\mathrm{coord}}$ are weights for different losses. All losses and their forward flows are shown in Fig.~\ref{fig:method}.

\subsection{Applications of FeatureNeRF}
\label{subsec:applications}

We demonstrate the effectiveness of the learned 3D semantic NeRF feature function $\mathbf{v}_{\mathrm{NeRF}}$ on various downstream applications: 2D/3D semantic keypoint transfer and object part segmentation. We deliberately apply simple, zero-shot methodologies on NeRF features, without any fine-tuning or post-process, to validate the proposed representations.

\noindent
\textbf{2D Tasks.} Given a single image $I$, we can render its NeRF feature map $F \in \mathbb{R}^{H \times W \times D}$ using Eq.~\ref{eq:feature_rendering}. Then we can render novel-view feature map $F^\prime$ from other viewpoints. For feature vectors $\mathbf{V}$ and $\mathbf{V}^\prime$ of two pixels from $F$ and $F^\prime$, we use cosine similarity to measure their distance in the feature space: $D(\mathbf{V}, \mathbf{V}^\prime) = \frac{\mathbf{V} \cdot \mathbf{V}^\prime}{\left\|\mathbf{V}\right\|_2 \cdot\left\|\mathbf{V}^\prime\right\|_2}$. For the part co-segmentation task, for each pixel feature $\mathbf{V}^\prime$ in $F^\prime$, we take the segmentation label of its closest pixel feature $\mathbf{V}$ in $F$ as the predicted segmentation label. For the keypoint transfer task, we adopt a similar process, for the feature $\mathbf{V}$ of each keypoint in $F$, we take the pixel location of its closest feature in $F^\prime$ as the predicted keypoint.

Since FeatureNeRF is a generalizable NeRF conditioned on the input image, it can be further applied to cross-instance tasks in addition to novel-view tasks for a single instance. Given images $I_1$ and $I_2$ of two instances, we can render their feature maps $F_1$ and $F_2$. Then we can resolve semantic correspondence tasks in a process similar to the novel-view tasks.

\noindent
\textbf{3D Tasks.} The FeatureNeRF model learned with only 2D observations can also be leveraged to 3D tasks. Given images $I_1$ and $I_2$ of two instances, we can construct two continuous 3D feature fields. For feature vectors $\mathbf{v}_1$ and  $\mathbf{v}_2$ of two 3D points  $\mathbf{x}_1$ and $\mathbf{x}_2$, we still utilize cosine similarity to measure their distance: $D(\mathbf{v}_1, \mathbf{v}_2) = \frac{\mathbf{v}_1 \cdot \mathbf{v}_2}{\left\|\mathbf{v}_1\right\|_2 \cdot\left\|\mathbf{v}_2 \right\|_2}$. Then we can resolve semantic correspondence tasks in a process similar to the 2D tasks.

\section{Experiments}
\label{sec:experiments}

\subsection{Experimental Setting}
\label{subsec:setting}

\noindent
\textbf{Datasets.} Our experiments are mainly conducted on 6 categories from the ShapeNet~\cite{chang2015shapenet} dataset: Chair, Car, Airplane, Table, Bottle and Motorcycle. We evaluate our model using annotations from from KeypointNet~\cite{you2020keypointnet}, ShapeNet part dataset~\cite{Yi16} and PartNet~\cite{Mo_2019_CVPR}. In addition, we train our model on the real-world CO3D~\cite{reizenstein2021common} dataset and evaluate its keypoint transfer performance on the Spair~\cite{Min2019SPair71kAL} dataset. For ShapeNet, we split each category into training (70\%), validation (10\%), and testing (20\%) splits. All shapes are normalized so that the longest edges of the bounding box are equal. For the training set, 50 random camera poses from the upper hemisphere are sampled. For validation and testing sets, 50 fixed camera poses are used. We employ blender~\cite{blender2018} to render RGB images, with a resolution of $128\times128$, the same as in PixelNeRF~\cite{yu2021pixelnerf}.  For PartNet, we use level-1 annotations. All annotations are in 3D and can be used for the evaluation of 3D tasks directly. For the evaluation of 2D tasks, we employ PyTorch3D~\cite{ravi2020pytorch3d} rasterizer to render 2D part segmentation labels and 2D keypoints.

\noindent
\textbf{Baselines.} We mainly compare FeatureNeRF quantitatively and qualitatively to two foundation models DINO~\cite{caron2021emerging} and Latent Diffusion~\cite{rombach2022high}. We extend PixelNeRF~\cite{yu2021pixelnerf} with the mechanisms mentioned in Sec.~\ref{subsec:descriptor} to make it possible for semantic understanding tasks and report its performance as ``NeRF feature" in all results. In addition, for the 2D co-segmentaion task, we re-implement a one-shot generalizable SemanticNeRF*~\cite{zhi2021place} by adding a semantic branch in PixelNeRF, and train it with the source segmentation labels.

\noindent
\textbf{Implementation Details.} Following PixelNeRF~\cite{yu2021pixelnerf}, we employ a ResNet-34 model pre-trained on ImageNet as the image encoder $f$. The batch size is 4 (objects) and 1024 rays per object. We train a single model for each object category for 500k steps. The weights for different losses are $\lambda_{\mathrm{distill}} = 0.25$ and $\lambda_{\mathrm{coord}} = 0.25$. The dimension of the internal NeRF feature is 512. MLP for NeRF feature $g_{\mathrm{NeRF}}$ is 4-layer, all other shallow MLPs for final outputs ($g_\sigma$, $g_{\mathbf{c}}$, $g_{\mathbf{v}}$ and $g_{\mathbf{x}}$) are one-layer.

\begin{figure*}[t!]
  \centering
  \vspace{-0.05in}
  \includegraphics[width=1.0\textwidth]{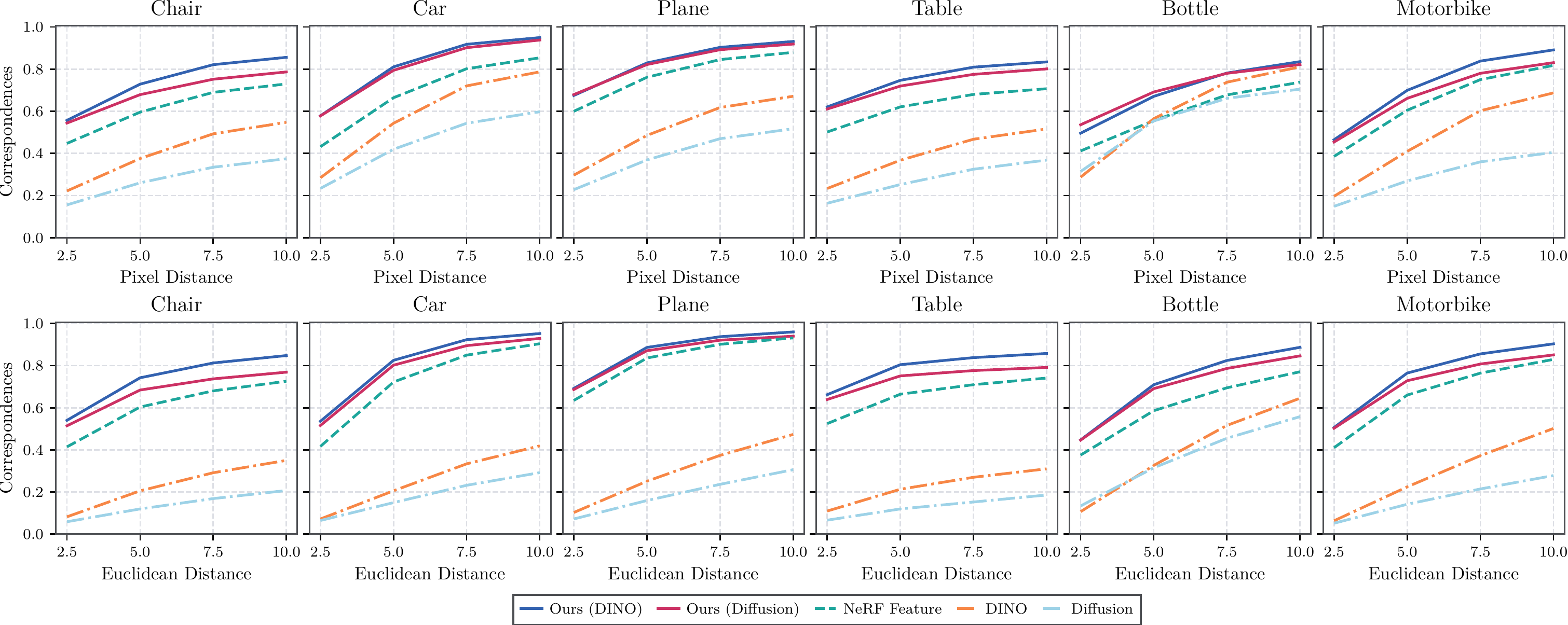}
  \vspace{-0.3in}
  \caption{\textbf{Correspondence accuracy for cross-instance semantic keypoints transfer}. The first row is for 2D keypoints transfer and the second row is for 3D. Our approach distilled with different features consistently outperforms baselines for all categories in both 2D and 3D domains. }
  \label{fig:kp_corre}
  \vspace{-0.05in}
\end{figure*}

\begin{figure*}[t!]
  \vspace{-0.1in}
  \centering
  \includegraphics[width=1.0\textwidth]{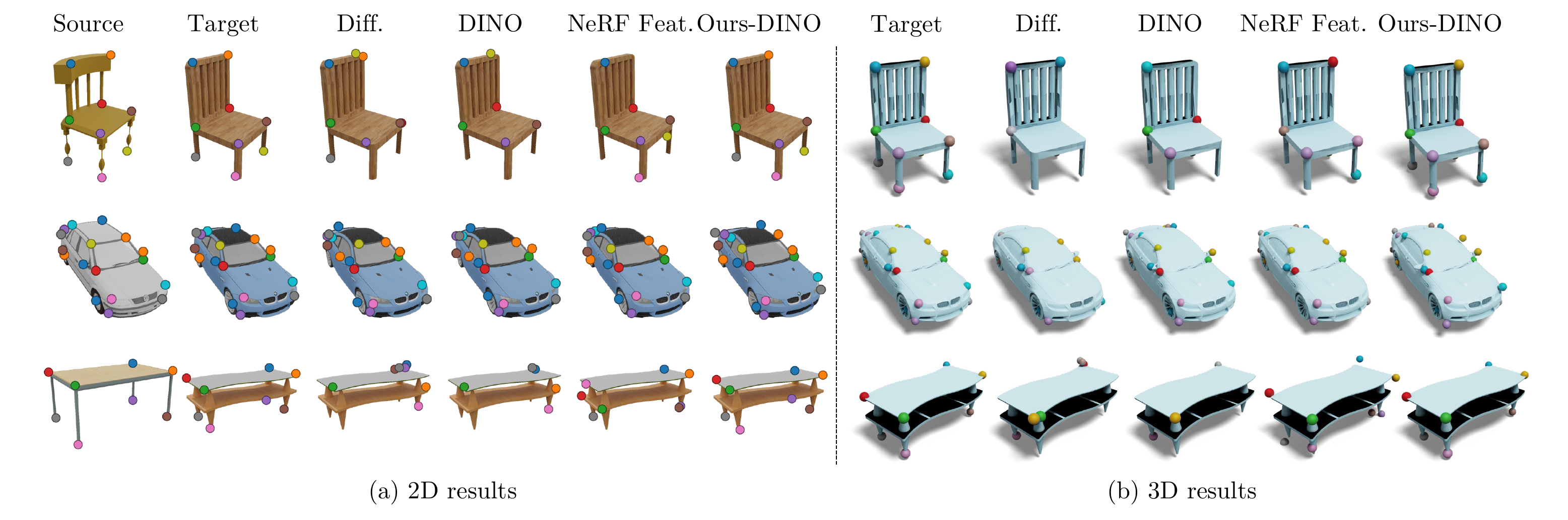}
  \vspace{-0.25in}
  \caption{\textbf{Qualitative results for cross-instance semantic keypoints transfer}. Both 2D (a) and 3D (b) results are presented here. Each row contains a source image with keypoints annotations and its pairwise transfer results. }
  \label{fig:kp_viz}
  \vspace{-0.1in}
\end{figure*}

\begin{figure*}[h]
  \centering
  \vspace{-0.15in}
  \includegraphics[width=1.0\textwidth]{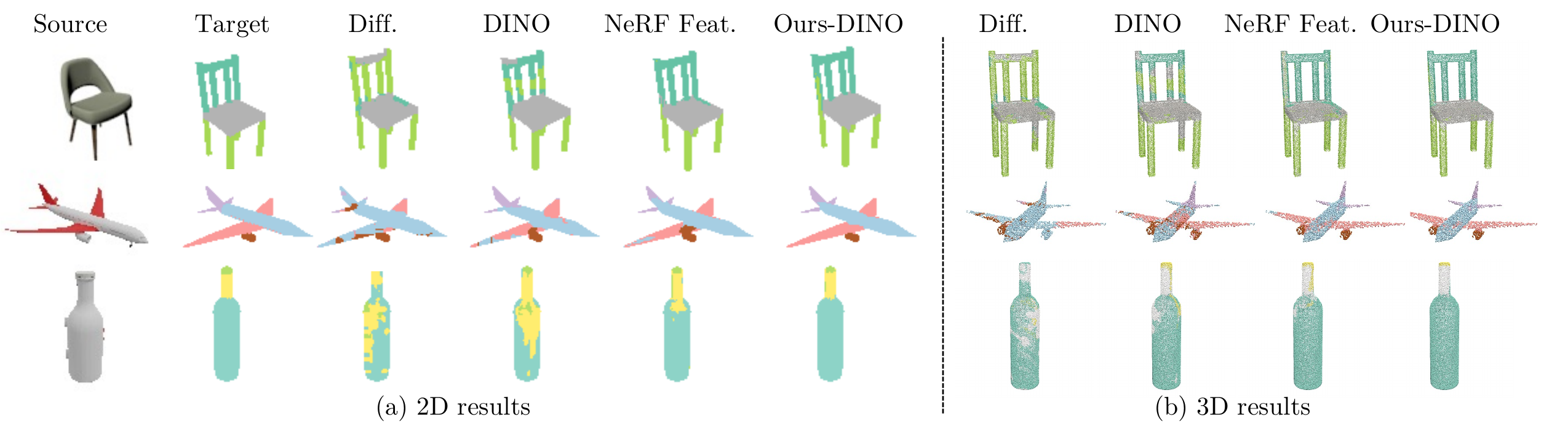}
  \vspace{-0.35in}
  \caption{\textbf{Qualitative results for cross-instance part segmentation label transfer}. Each row contains a source image and its 2D/3D transfer results. After distilling, FeatureNeRF learns richer semantic information, produces better boundaries and preserves details like small parts. Note that the segmentation label for the source instance is omitted. }
  \label{fig:part_seg}
\vspace{-0.05in}
\end{figure*}

\begin{figure*}[h]
  \centering
  \vspace{-0.1in}
  \includegraphics[width=0.97\textwidth]{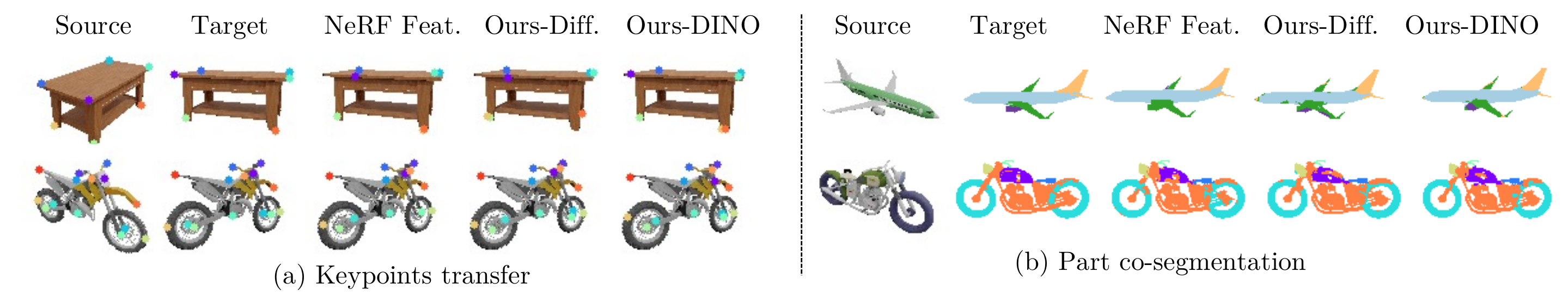}
  \vspace{-0.2in}
  \caption{\textbf{Qualitative results of novel-view keypoints transfer and part co-segmentation.} FeatureNeRF learns a 3D representation from 2D observations, which makes it possible to synthesize feature maps from other viewpoints and transfer keypoints and segmentation labels to them. }
  \label{fig:nv_results}
  \vspace{-0.1in}
\end{figure*}

\noindent
\textbf{Teacher Networks.}
We employ DINO~\cite{caron2021emerging} and Latent Diffusion~\cite{rombach2022high} as teacher networks, which are pre-trained and publicly available. The patch size of DINO is 8. The final feature map has dimension $32 \times 32 \times 384$. For Latent Diffusion, we extract the outputs of the 4th layer in U-Net as teacher features. The language prompts used to condition denoising modules are class names (e.g. ``Chair", ``Car") from ShapeNet. The final feature map has dimension $128 \times 128 \times 960$. The weights of teacher networks are fixed during distillation.

\subsection{2D Semantic Understanding Tasks}
\label{subsec:2d_task}

As mentioned in Sec.~\ref{subsec:applications}, we evaluate FeatureNeRF on tasks of 2D keypoints transfer and part segmentation labels transfer under two settings: (i) \textit{Cross-instance}: given images of two instances, we transfer keypoints/segmentation labels from the source image to the target image and compare the transferred labels in the target image with the ground truth. (ii) \textit{Novel-view}: given a single image of an instance, we render a novel-view feature map, transfer labels from the source viewpoint to the target viewpoint and compare the transferred labels in the target viewpoint with the ground truth.

\noindent
\textbf{Metrics.} For the task of 2D keypoints transfer, we report the percentage of predicted keypoints whose distances from their corresponding ground truths are below thresholds of $(2.5, 5.0, 7.5, 10.0)$ pixels in the target image. We denote this percentage as \textit{Correspondence Accuracy} in the following. For the task of part segmentation labels transfer, we calculate the mean intersection over union (\textit{mIoU}) over every part category for each object class. For both settings, we randomly generate 1000 combinations per category. While generating, we make sure that the source viewpoints/instances and target viewpoints/instances have intersecting keypoints/segmentation labels.

\noindent
\textbf{2D Keypoints Transfer.} We report correspondence accuracy of 2D cross-instance keypoints transfer in the first row of Fig.~\ref{fig:kp_corre}. We show that for all 6 categories, keypoints transferred via our proposed method are more accurate than baselines. The performances of distilling diffusion and DINO features are similar. Using NeRF feature without distilling foundation models (denoted as ``NeRF feature" in the figure) also achieves a reasonable performance compared to 2D foundation model baselines. Fig.~\ref{fig:kp_viz} (a) shows the qualitative results for 2D keypoints transfer. It can be seen that despite various appearances and structures of instances, our method can successfully transfer keypoints based on semantic understanding, while baselines often fails.

\begin{table*}[h!]
  \centering
  \tablestyle{5pt}{1.1}
  \vspace{-0.05in}
  \begin{tabular}{l|cccccc|cccccc}
    & \multicolumn{6}{c|}{2D part co-segmentation} & \multicolumn{6}{c}{3D part co-segmentation} \\
    & Chair  & Car & Plane & Table & Bottle & Motorbike & Chair  & Car & Plane & Table & Bottle & Motorbike \\
    \shline
    SemanticNeRF$^*$ & 50.32 & 34.61 & 56.19  & 54.87 & 51.58 & 27.62 & \multicolumn{6}{c}{-} \\
    Diffusion Baseline & 41.59 & 42.46 & 44.60 & 57.77 & 41.72 & 25.50 & 33.60 & 24.74 & 30.45 & 44.34 & 53.81 & 20.01   \\
    DINO Baseline & 62.43 & 54.59 & 57.81 & 64.01 & 59.73 & 37.05 & 52.67 & 29.72 & 40.16 & 49.98 & 70.79 & 29.25 \\
    NeRF Feature & 72.02 & 58.57 & 72.57 & 70.41 & 55.87 & 44.00 & 65.23 & 59.19 & 71.23 & 61.64 & 66.63 & 44.88 \\
    Ours (Diffusion) & 65.39 & 63.02 & 72.95 & 70.59 & 53.91 & 44.84 & 63.65 & 61.27 & 73.41 & 63.93 & 66.91 & 49.05 \\
    Ours (DINO) & \textbf{76.55} & \textbf{66.85} & \textbf{74.60} & \textbf{74.06} & \textbf{61.13} & \textbf{49.56} &  \textbf{73.85} & \textbf{64.99} & \textbf{74.20} & \textbf{66.52} & \textbf{72.33} & \textbf{52.56} \\
  \end{tabular}
  \vspace{-0.15in}
  \caption{\textbf{Cross-instance part segmentation label transfer results}. We report mIoU of part co-segmentation for each category. The left part is for 2D and the right is for 3D. By distilling features to the 3D space, our proposed representation contains richer semantic information which is apt for this co-segmentation task.}
  \label{tab:part_seg}
  \vspace{-0.05in}
\end{table*}

FeatureNeRF learns a 3D representation from 2D observations, which allows to synthesize novel-view feature maps for different viewpoints and performs keypoints transfer on top of it. Fig.~\ref{fig:nv_results} (a) shows qualitative results of novel-view keypoints transfer. Since 2D foundation models can not synthesize novel-view feature maps, we do not apply them to this novel-view setting.

\noindent
\textbf{2D Part Segmentation Label Transfer.}  We report mIoU results of 2D cross-instance part segmentation label transfer in the left part of Tab.~\ref{tab:part_seg}. FeatureNeRF distilled with DINO features significantly outperforms other approaches for all 6 categories. However, the performance of FeatureNeRF distilled with diffusion features is not as good as the DINO one. This is possible since the keypoints transfer task may focus on predicting accurate locations of sparse pixels that contain the richest semantic information, but part co-segmentation requires denser correspondences. Qualitative results are shown Fig.~\ref{fig:part_seg} (a). We can find that NeRF feature often produces unexpected artifacts. We attribute this phenomenon to NeRF feature's training only relying on RGB information. In contrast, by distilling pre-trained features, our method produces better boundaries and preserves details like small parts. Note that the segmentation label of the source object is also required during the transfer process, which is omitted in the figure.

We also perform novel-view part co-segmentation and report both quantitative and qualitative results in Tab.~\ref{tab:nv_2d_part} and Fig.~\ref{fig:nv_results} (b) respectively. The rendered feature maps from other viewpoints still exhibit promising performance for co-segmentation, which proves that FeatureNeRF learns a 3D-consistent feature representation.

\begin{figure}[t!]
  \centering
  \includegraphics[width=0.75\linewidth]{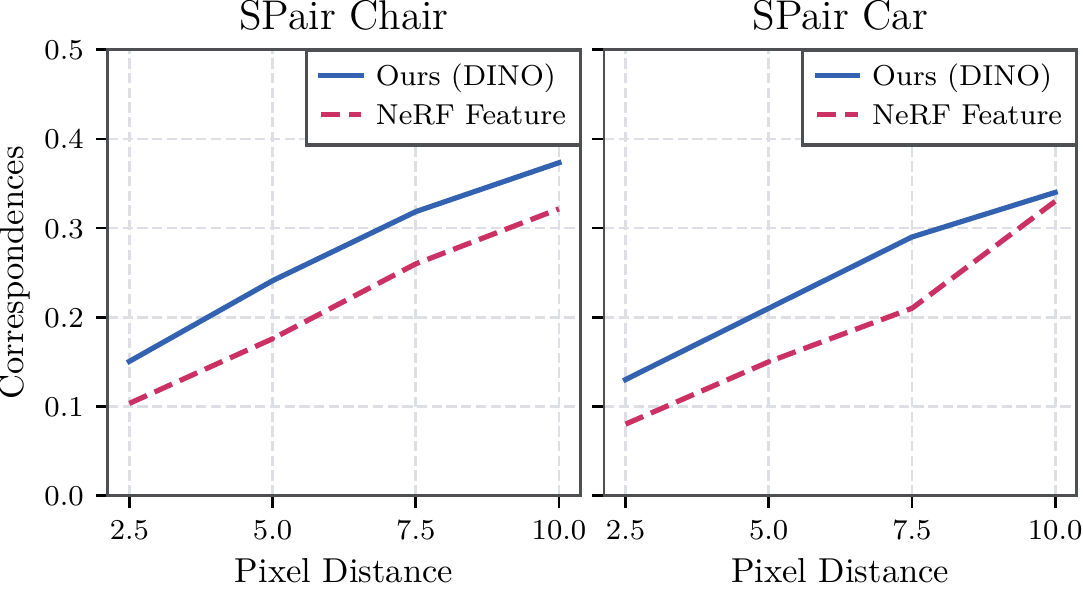}
  \vspace{-0.15in}
  \caption{\textbf{Keypoints transfer on SPair}. Our method outperforms NeRF feature on real-world images. }
  \label{fig:2d_kp_real}
  \vspace{-0.1in}
\end{figure}

\begin{figure}[t!]
  \centering
  \vspace{-0.1in}
  \includegraphics[width=0.95\linewidth]{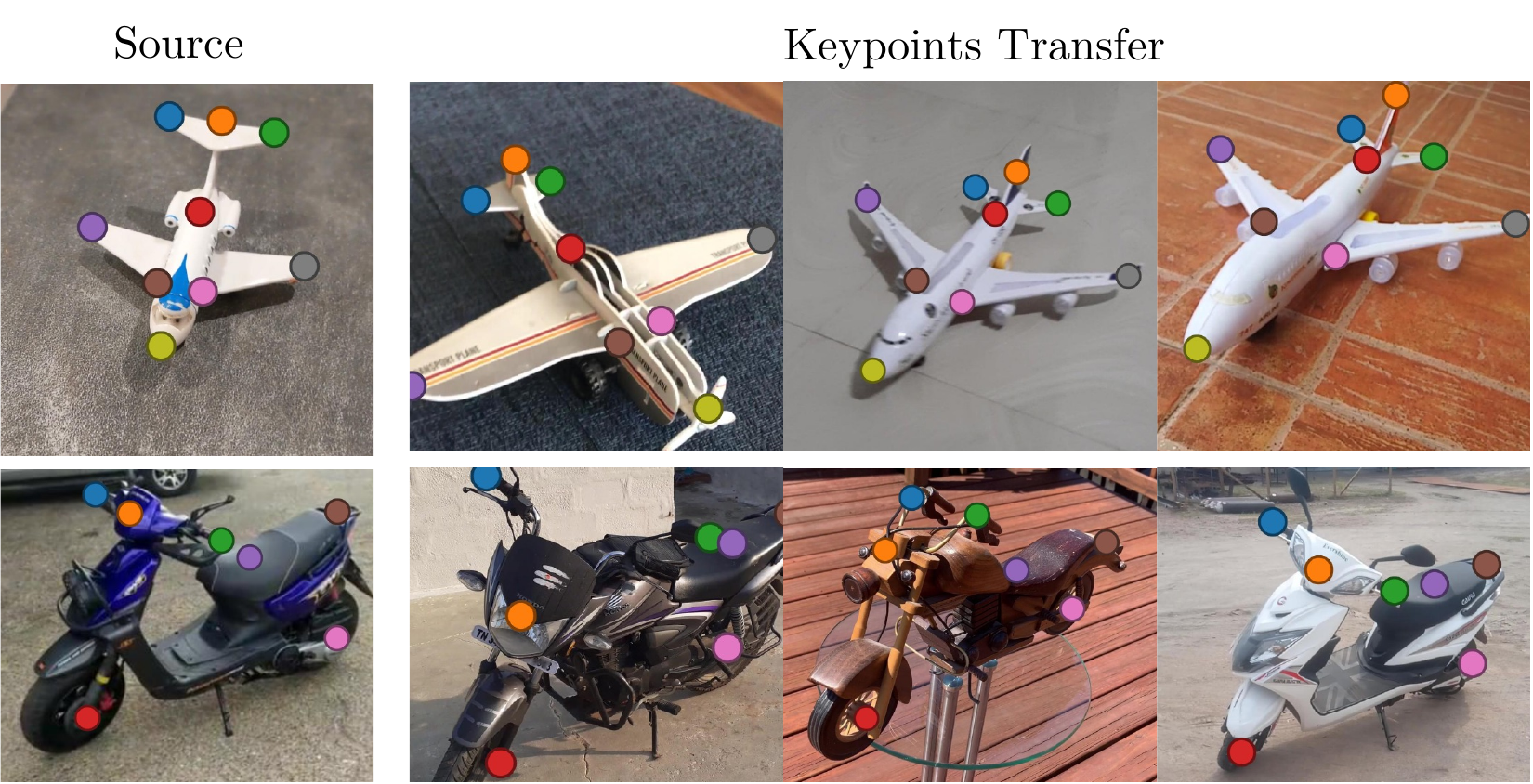}
  \vspace{-0.1in}
  \caption{\textbf{Keypoints transfer on CO3D}. After training on CO3D, our method can accurately transfer keypoints with in-the-wild image inputs }
  \label{fig:real_car}
  \vspace{-0.15in}
\end{figure}

\noindent
\textbf{Real-world Experiments.}
To show the generalizability to real-world images, we fine-tune our model on CO3D datasets and evaluate its keypoints transfer performance on two categories from the SPair dataset. The results in Fig.~\ref{fig:2d_kp_real} confirm the effectiveness of our approach over NeRF feature on real images.
The qualitative results on CO3D are shown in Fig.~\ref{fig:real_car}. Even with in-the-wild images, our method can still transfer keypoints accurately. Note that pre-process steps (e.g. segment and normalize the foreground object) are required.

\subsection{3D Semantic Understanding Tasks}
\label{subsec:3d_task}
\vspace{-0.05in}

We further validate our proposed method on 3D semantic understanding tasks, which aims to find semantic correspondences between two sets of 3D points based on image observations. We only evaluate under the cross-instance setting for 3D tasks.

\noindent
\textbf{Metrics.} The metrics for 3D tasks are 3D versions of their counterparts in 2D tasks. For the keypoints transfer, we utilize Euclidean distance instead of 2D pixel distance with the threshold of $(0.025, 0.05, 0.075, 0.1)$ in the normalized space. For the part co-segmentation, we calculate mIoU based on point clouds instead of pixels. We randomly generate 1000 combinations per category for the evaluation.

\noindent
\textbf{3D Keypoints Transfer.} For each 3D keypoint from the source shape, we try to find its corresponding point from a set of sampled candidate points from the target shape. For DINO and diffusion baselines, we simply project the 3D points onto 2D feature maps and utilize the interpolated feature vectors for matching. As those features are not 3D-aware, it is expected that their performance will be suboptimal. The quantitative results are shown in the second row of Fig~\ref{fig:kp_corre}. We can find that our method has a larger advantage in 3D tasks, which can also be observed from the qualitative results in Fig~\ref{fig:kp_viz}. Our learned 3D representation contains accurate semantic information and thus are able to transfer 3D keypoints in a more accurate way.

\begin{table}[t!]
  \centering
  \tablestyle{4pt}{1.1}
  \begin{tabular}{l|cccccc}
    & Chair & Car & Plane & Table & Bottle & Motorbike \\
    \shline
    NeRF Feat. & 80.73 & 75.10 & 69.45 & 84.65 & 87.14 & 67.49 \\
    Ours (Diff.) &  73.45 & 71.16 & 59.16 & 83.91 & \textbf{88.01} & 64.28 \\
    Ours (DINO) & \textbf{81.93} & \textbf{76.50} & \textbf{71.57} & \textbf{87.97} & 87.89 & \textbf{68.22} \\
  \end{tabular}
  \vspace{-0.1in}
  \caption{\textbf{Novel-view 2D part segmentation label transfer results}.  The rendered feature maps from other viewpoints still exhibit promising performance for co-segmentation, which proves that FeatureNeRF learns a 3D-consistent feature representation. }
  \label{tab:nv_2d_part}
  \vspace{-0.1in}
\end{table}

\noindent
\textbf{3D Part Segmentation Label Transfer.} Similar to keypoints transfer, we try to propagate segmentation labels from the source point cloud to the target point cloud based on image observations. Note that the point cloud is only used as a query for segmentation, our method constructs a semantic feature volume for the whole 3D space and allows predicting feature vectors for any given 3D points. The right part of Tab~\ref{tab:part_seg} reports mIoUs for all methods. FeatureNeRF consistently outperforms all baselines in all categories, suggesting that distilling feature to 3D-aware representation leads to improved 3D co-segmentation performance. Fig.~\ref{fig:part_seg} presents qualitative results. Note that even for occluded areas in 2D images, FeatureNeRF can transfer its labels correctly in 3D space.

\subsection{Ablation Study}
\label{subsec:ablation}
\vspace{-0.05in}

We mainly ablate the coordinate loss $\mathcal{L}_{\mathrm{coord}}$ and the use of internal NeRF features $\mathbf{v}_{\mathrm{NeRF}}$ (instead of final output feature vector $\mathbf{v}$) for semantic understanding. We conduct experiments of 2D parts co-segmentation on the Chair and Plane classes. The mIoU results are reported in Tab.~\ref{tab:ablation}. We can see that both design choices boost performance.

In addition, we compare the performance of novel-view synthesis with PixelNeRF~\cite{yu2021pixelnerf} to see if the proposed feature distillation technique hurts the synthesis ability. Both FeatureNeRF and PixelNeRF are trained on our rendered dataset with the same parameters. From Tab.~\ref{tab:nv_synthesis}, we see that our method achieves comparable performance with PixelNeRF on novel-view synthesis.

\begin{table}[t]
  \centering
  \tablestyle{3pt}{1.1}
  \begin{tabular}{l|cc}
    Method & Chair & Plane \\
    \shline
    w/o $\mathcal{L}_{\mathrm{coord}}$ & 75.39 & 72.83 \\
    w/o internal features & 74.62 & 72.29 \\
    full model & \textbf{76.55} & \textbf{74.60} \\
  \end{tabular}
  \vspace{-0.1in}
  \caption{\textbf{Ablation study}. The coordinate loss and the use of internal NeRF features both boost performance. }
  \label{tab:ablation}
  \vspace{-0.15in}
\end{table}

\begin{table}[t]
  \centering
  \tablestyle{3pt}{1.1}
  \begin{tabular}{ll|cccccc}
    & & Chair & Car & Plane & Table & Bottle & Motorbike \\
    \shline
    \multirow{3}{*}{PSNR} & PixelNeRF~\cite{yu2021pixelnerf} & 23.29 & 22.86 & 24.46 & 25.61 &
\textbf{26.04}  & 20.36 \\
    & Ours (Diff.) & \textbf{23.36} & \textbf{23.09} & 24.39 & 25.42 & 25.75 & \textbf{20.64} \\
    & Ours (DINO) & 23.20 & 22.92 & \textbf{24.49} & \textbf{25.69} & 26.01 & 20.59 \\
    \hline
    \multirow{3}{*}{SSIM} & PixelNeRF~\cite{yu2021pixelnerf} & 0.92 & 0.91 & \textbf{0.93} & 0.89 & 0.89 & 0.80 \\
    & Ours (Diff.) & \textbf{0.92}  & 0.91 & 0.91 & 0.87 & 0.89 & \textbf{0.81} \\
    & Ours (DINO) & 0.91 & \textbf{0.91} & 0.91  & \textbf{0.89} &  \textbf{0.90} & 0.80\\
  \end{tabular}
    \vspace{-0.1in}
  \caption{\textbf{Novel-view synthesis results}. The proposed distillation process does not hurt the performance of novel-view synthesis, our method achieves comparable performance with PixelNeRF on novel-view synthesis. }
  \label{tab:nv_synthesis}
  \vspace{-0.15in}
\end{table}

\vspace{-0.05in}
\section{Conclusion}
\label{sec:conclusion}
\vspace{-0.05in}

In this paper, we present FeatureNeRF, a unified framework for learning generalizable NeRFs from distilling 2D vision foundation models. FeatureNeRF explores the use of internal NeRF features as 3D visual descriptors and distills knowledge from foundation models into 3D space via neural rendering. Given a single image, FeatureNeRF allows to predict a 3D semantic feature volume, which can be leveraged for downstream tasks. Specifically, we demonstrate the effectiveness of FeatureNeRF on the tasks of 2D/3D keypoints transfer and part co-segmentation from a single image. We envision exploring more applications such as object editing with FeatureNeRF in future work.

{\small
\bibliographystyle{ieee_fullname}
\bibliography{egbib}
}

\begin{figure*}[t!]
  \centering
  \includegraphics[width=1.0\textwidth]{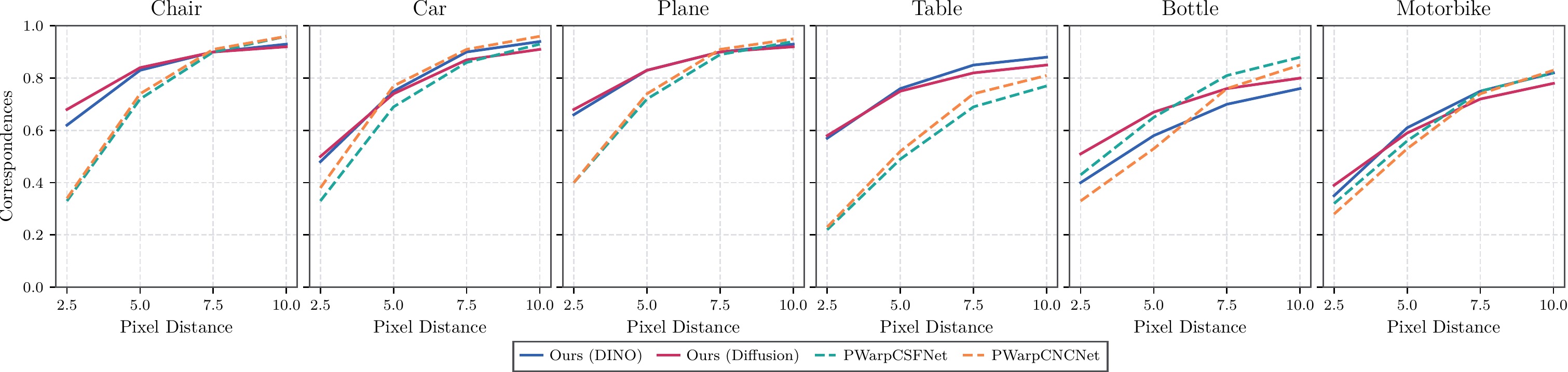}
  \vspace{-0.25in}
  \caption{\textbf{Correspondence accuracy for cross-instance semantic keypoints transfer}.  We additionally compare FeatureNeRF to an off-the-shelf semantic matching approach PWarpC~\cite{truong2022probabilistic}. Our method achieves comparable performances under thresholds of larger pixel distances but outperforms them under smaller pixel distances.}
  \label{fig:2d_kp_supp}
  \vspace{-0.05in}
\end{figure*}

\newpage

\appendix

In the appendix, we provide additional quantitative and qualitative results, including new editing applications.

\section{2D Keypoints Transfer}

For the task of 2D semantic keypoints transfer, we additionally compare FeatureNeRF to an off-the-shelf semantic matching approach PWarpC~\cite{truong2022probabilistic}, which is weakly-supervised and achieves state-of-the-art performances on 2D semantic matching tasks. We directly deploy the pre-trained model to our rendered images. We employ two variants PWarpCSFNet and PWarpCNCNet, and report the results in Fig~\ref{fig:2d_kp_supp}. It can be seen that our method achieves comparable performances under thresholds of larger pixel distances but outperforms them under thresholds of smaller pixel distances.

\section{Editing Applications}

The learned FeatureNeRF model can also be leveraged to editing applications. Here, we take the 3D part texture swapping as an example. Given a source image and its part segmentation label, we can construct a 3D feature volume for the source image. Then, for a target image, we also construct its 3D feature volume and transfer the segmentation label from the source image to it. When rendering the target image, for a 3D point $\mathbf{x}_{\mathrm{tgt}}$ that belongs to the part of interests (e.g. chair back) in the target feature volume, we find its closet point in the source feature volume:
$$
\mathbf{x}_{\mathrm{closet}} = \argmin_{\mathbf{x}_{\mathrm{src}}} \| v_{\mathrm{NeRF}}\left(\mathbf{x}_{\mathrm{tgt}}\right) - v_{\mathrm{NeRF}}\left(\mathbf{x}_{\mathrm{src}}\right) \|_2,
$$
where $\mathbf{x}_{\mathrm{src}} \in \mathbf{X}_{\mathrm{src}}$ and $\mathbf{X}_{\mathrm{src}}$ is the set of sampled points in the source feature volume. Finally, we use the color of the closet point $\mathbf{x}_{\mathrm{closet}}$ to replace the original color of $\mathbf{x}_{\mathrm{tgt}}$ for the rendering.

\begin{figure}[t!]
  \centering
  \includegraphics[width=1.0\linewidth]{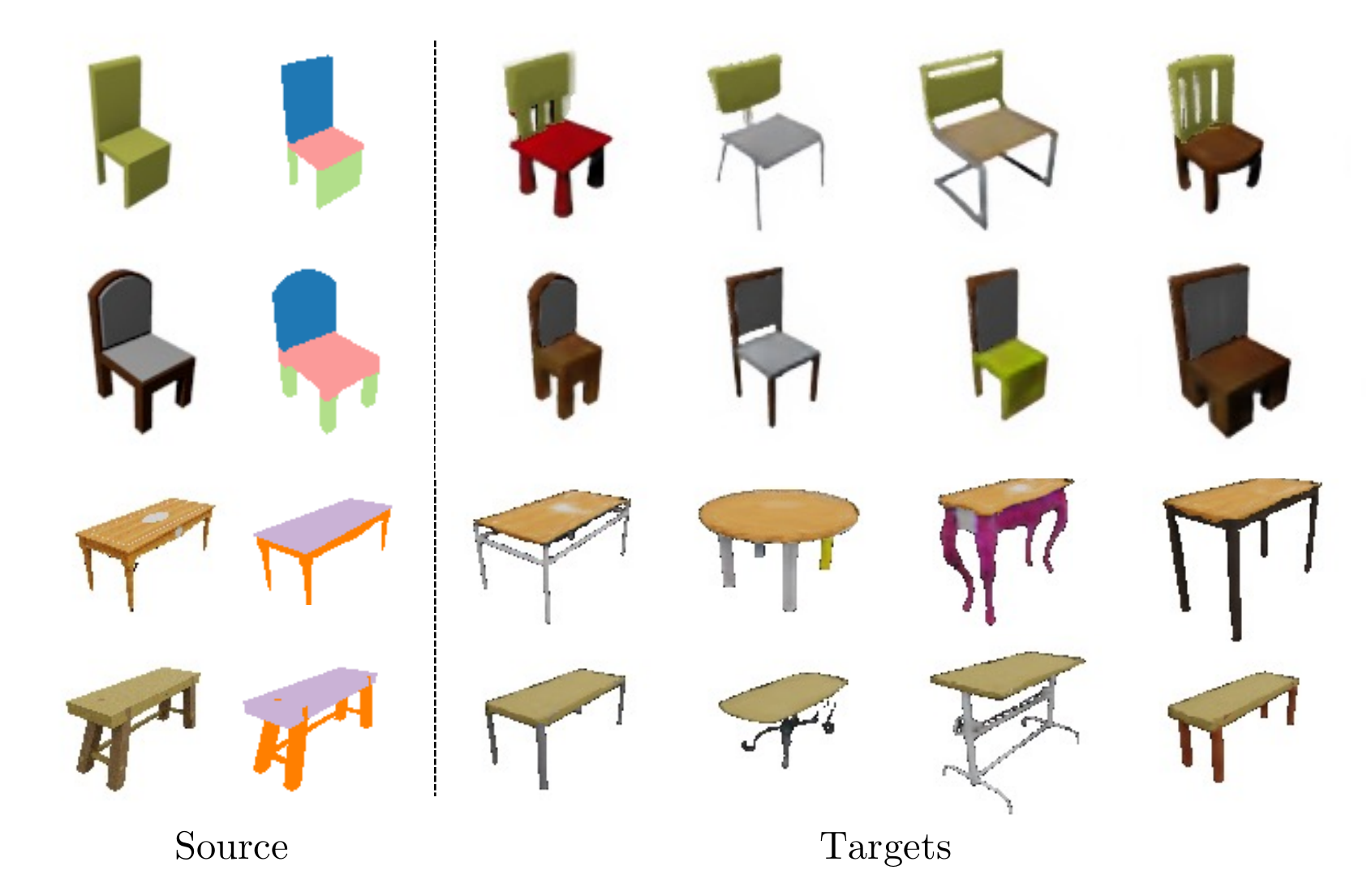}
  \vspace{-0.3in}
  \caption{\textbf{Editing application: 3D part texture swapping}. The part textures from source images are successfully transferred to the target images. Note that the edited target objects can also be rendered from novel views, please see our video for 3D visualizations.}
  \label{fig:supp_editing}
  \vspace{-0.15in}
\end{figure}

The results are shown in Fig.~\ref{fig:supp_editing}. We can find that the part textures from source images are successfully transferred to the target images. Besides, the details are also well preserved (the brown boundary from the source image at the second row still exists in the target images). Note that the edited target instances can also be rendered from novel views, please see our video for 3D visualizations.

\begin{figure*}[t!]
  \centering
  \includegraphics[width=1.0\linewidth]{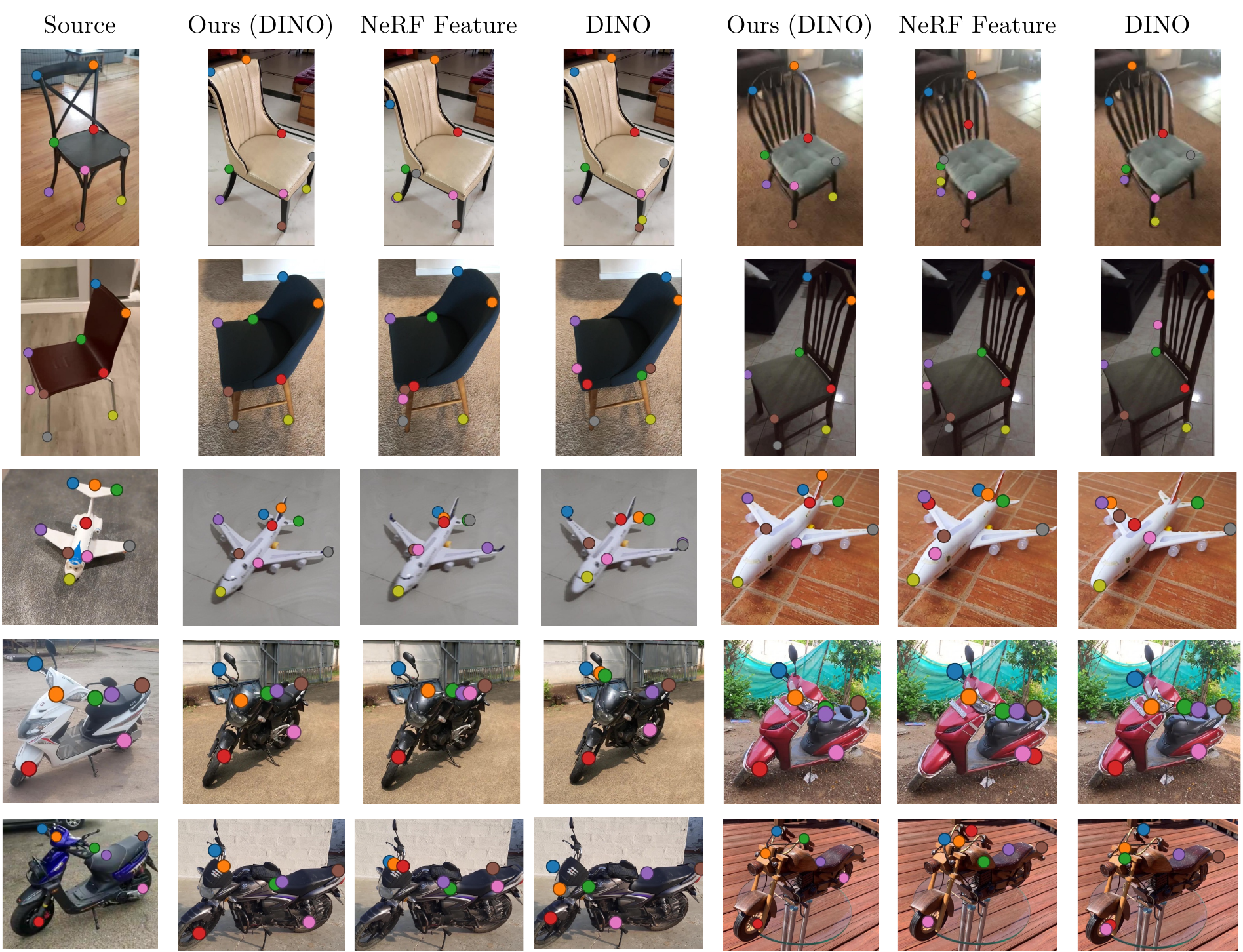}
  \vspace{-0.1in}
  \caption{\textbf{Keypoints transfer on CO3D}. }
  \label{fig:supp_real}
  \vspace{-0.1in}
\end{figure*}

\begin{figure*}[t!]
  \centering
  \includegraphics[width=1.0\linewidth]{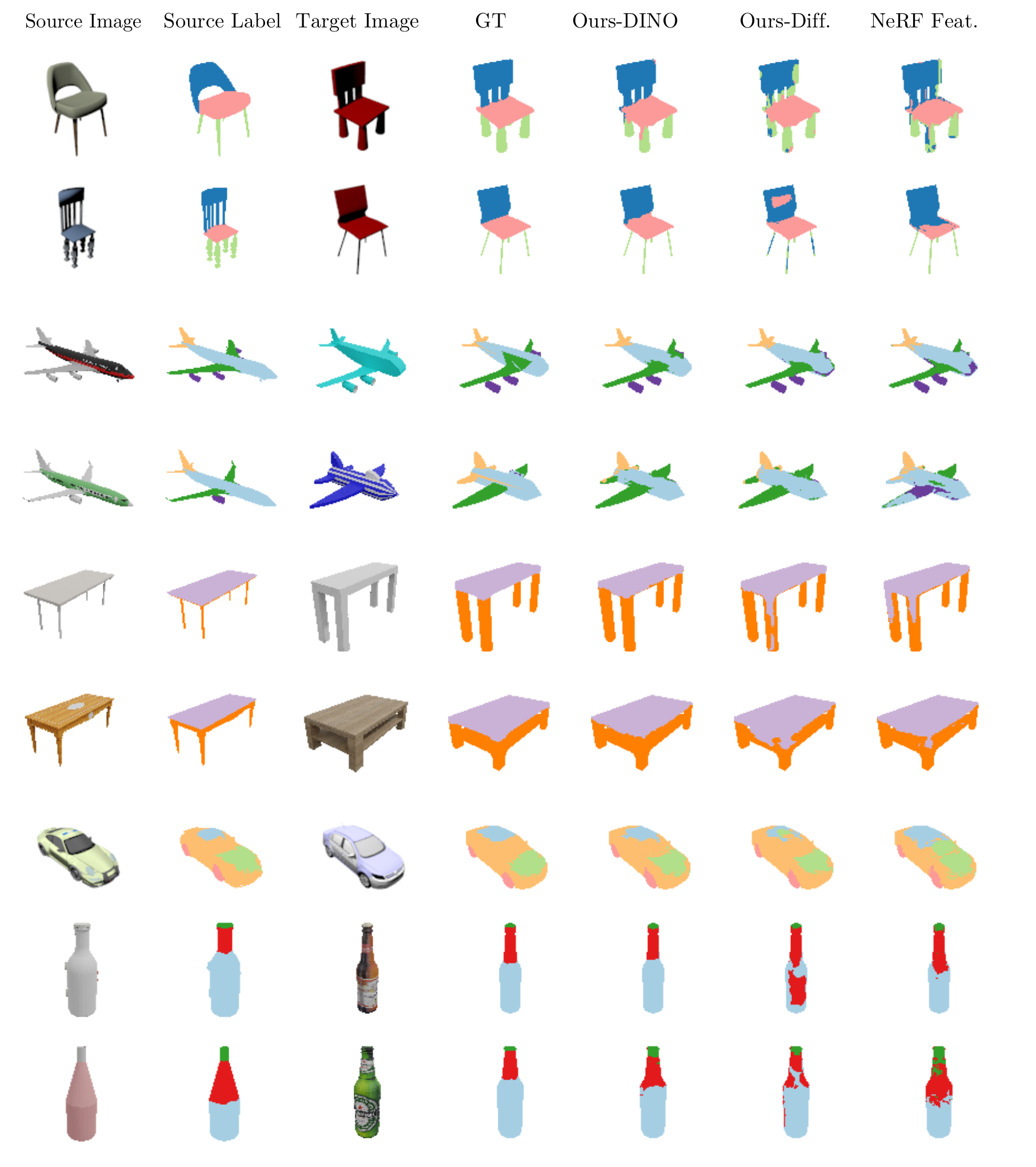}
  \vspace{-0.1in}
  \caption{\textbf{Qualitative results for cross-instance part co-segmentation}. }
  \label{fig:supp_seg}
  \vspace{-0.1in}
\end{figure*}

\begin{figure*}[t!]
  \centering
  \includegraphics[width=1.0\linewidth]{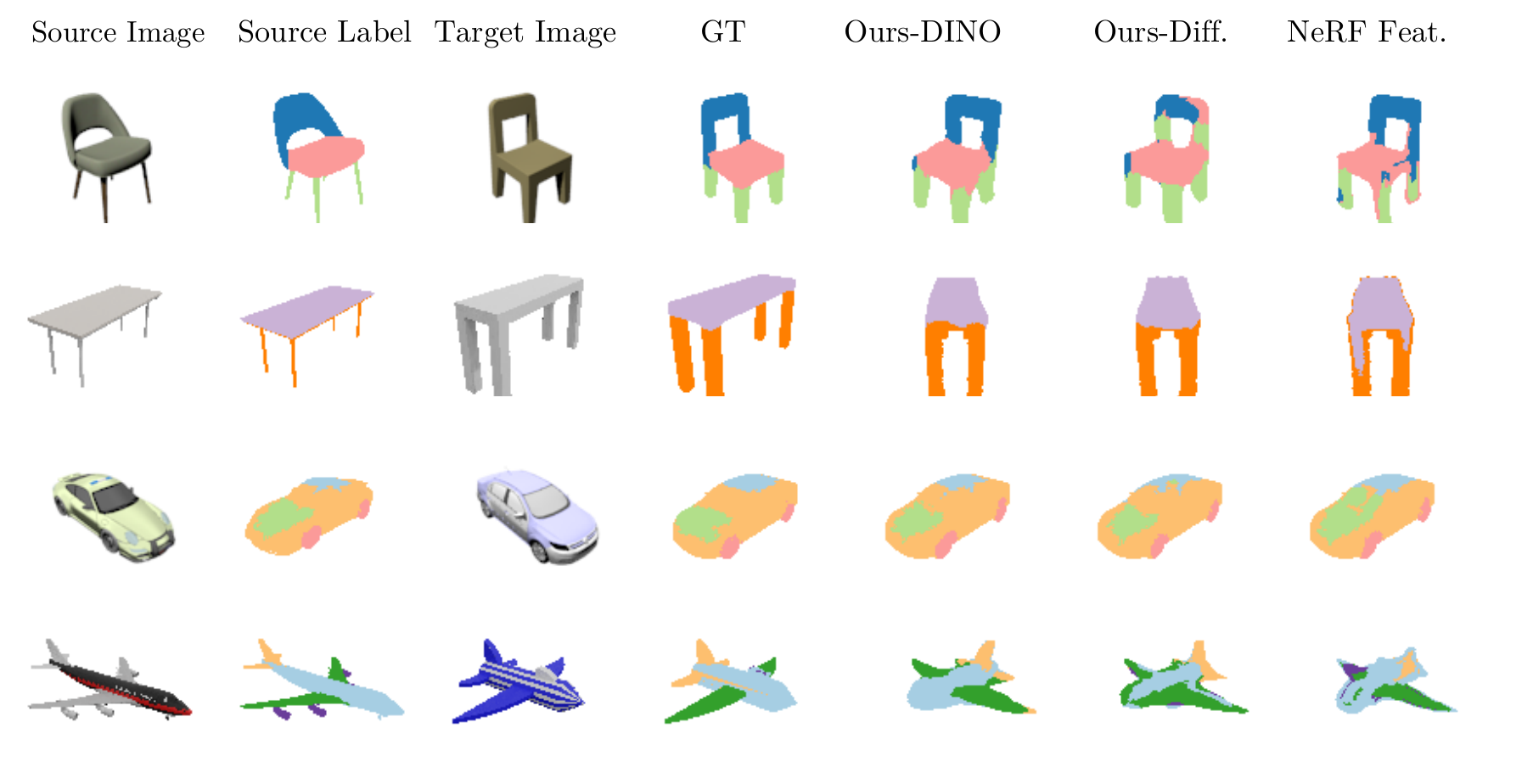}
  \vspace{-0.1in}
  \caption{\textbf{Qualitative results for novel-view cross-instance part co-segmentation}. }
  \label{fig:supp_nv_seg}
  \vspace{-0.1in}
\end{figure*}

\begin{figure*}[t!]
  \centering
  \includegraphics[width=0.75\linewidth]{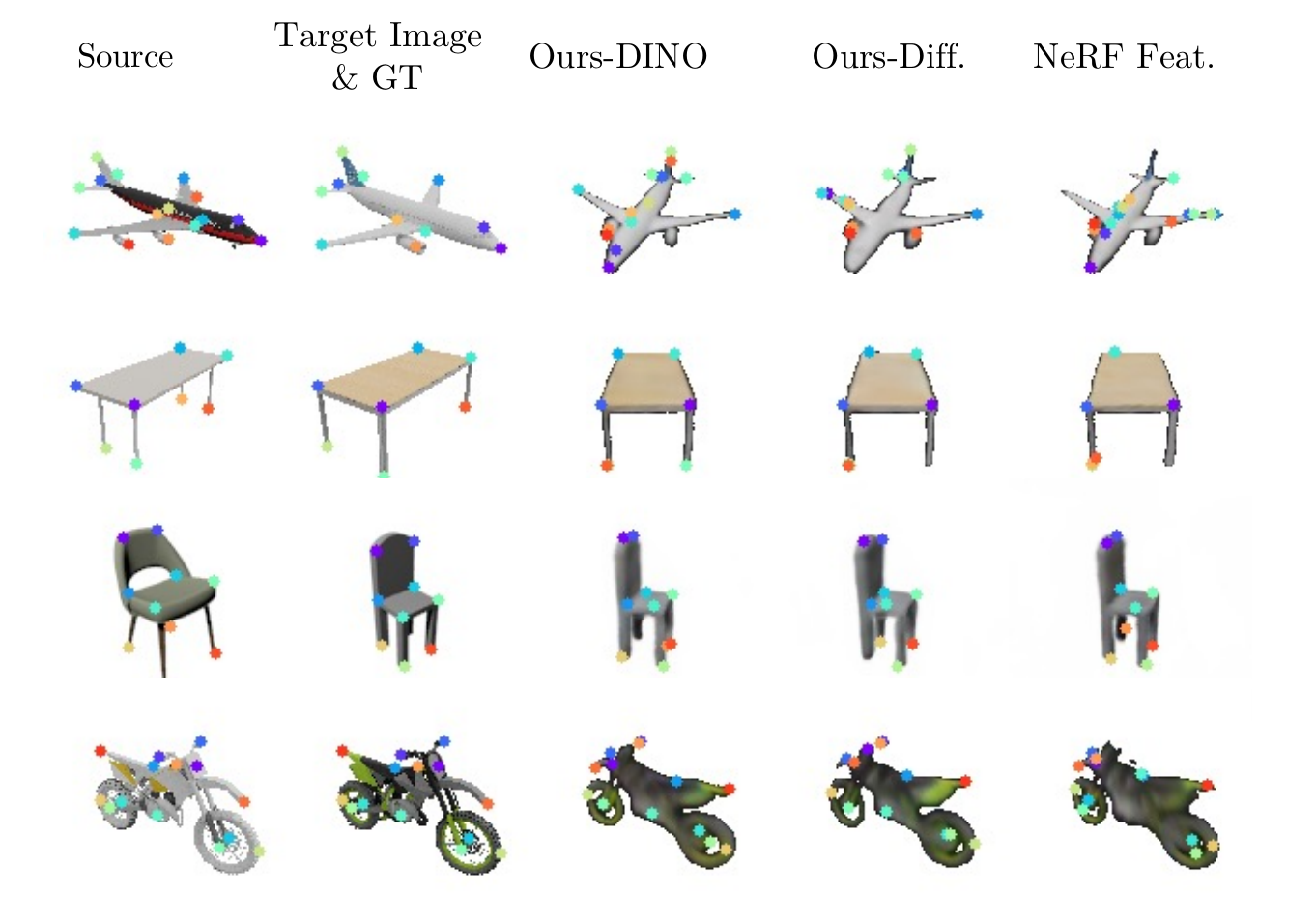}
  \vspace{-0.1in}
  \caption{\textbf{Qualitative results for novel-view cross-instance keypoints transfer}. }
  \label{fig:supp_nv_kp}
  \vspace{-0.1in}
\end{figure*}

\begin{figure*}[t!]
  \centering
  \includegraphics[width=0.75\linewidth]{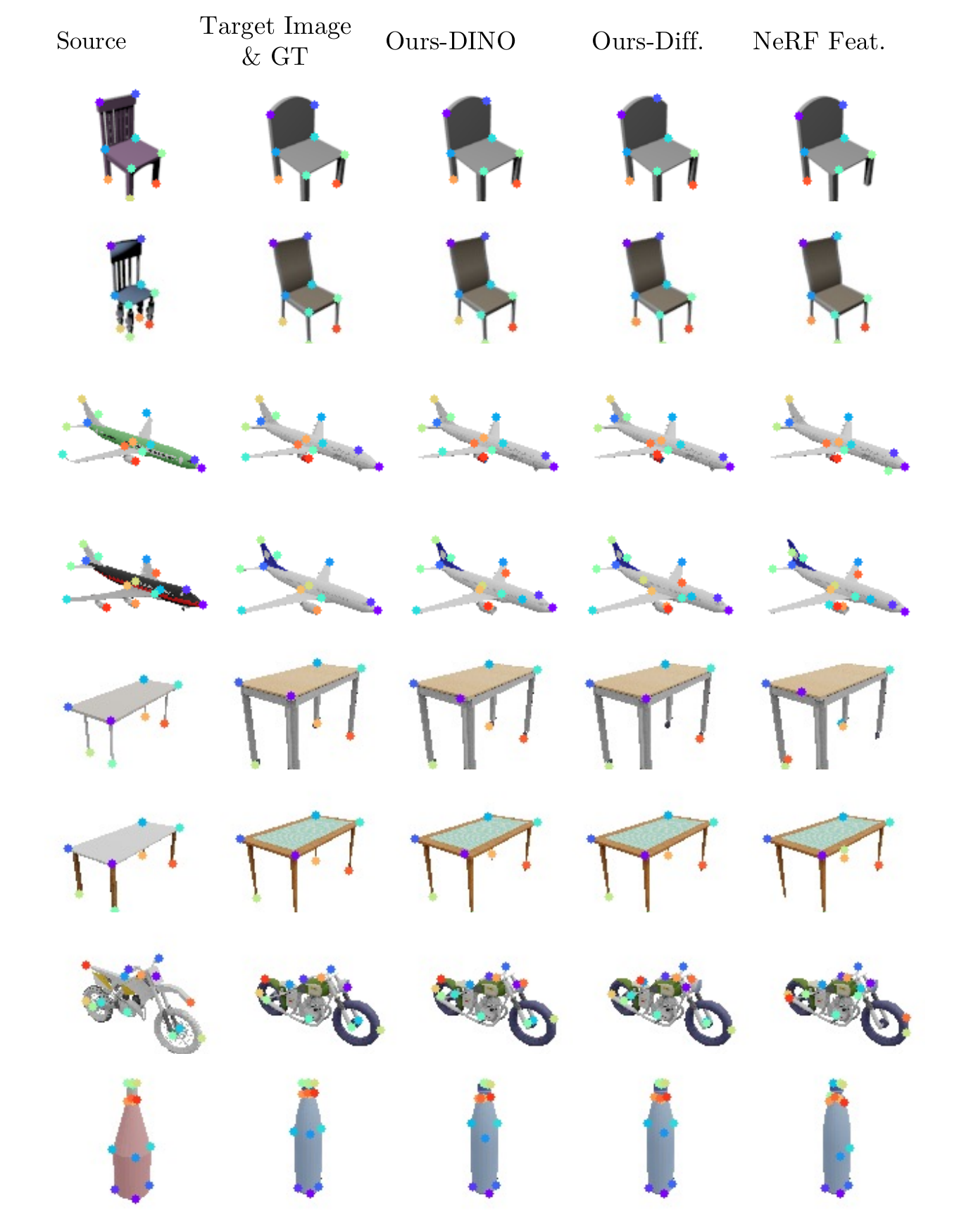}
  \vspace{-0.1in}
  \caption{\textbf{Qualitative results for cross-instance keypoints transfer}. }
  \label{fig:supp_kp}
  \vspace{-0.1in}
\end{figure*}

\section{Qualitative Results}

We present additional qualitative results for CO3D keypoint transfer in Fig~\ref{fig:supp_real}, cross-instance part co-segmentation in Fig~\ref{fig:supp_seg} and Fig~\ref{fig:supp_nv_seg}, and cross-instance keypoint transfer in Fig~\ref{fig:supp_kp} and Fig~\ref{fig:supp_nv_kp}.

\end{document}